\documentclass[letterpaper, 10 pt, conference]{ieeeconf}  

\IEEEoverridecommandlockouts                              

\usepackage{color}
\usepackage{colortbl}
\usepackage{xcolor}
\usepackage{graphicx}
\usepackage{subcaption}
\usepackage{gensymb}

\usepackage{amsmath}
\usepackage{hyperref}
\usepackage{multirow}
\usepackage{array}
\usepackage{booktabs}

\definecolor{todo-red}{RGB}{200,12,12}
\definecolor{green4}{RGB}{0,128,0}

\title{\LARGE \bf
Long-term Large-scale Mapping and Localization Using maplab
}

\author{Marcin Dymczyk, Marius Fehr, Thomas Schneider and Roland Siegwart
\thanks{All authors are with Autonomous Systems Lab, ETH Z\"urich, Switzerland.
        {\tt\small \{marcin.dymczyk, marius.fehr, thomas.schneider\}@mavt.ethz.ch, rsiegwart@ethz.ch}}%
}

\makeatletter
\renewcommand\@makefntext[1]{\leftskip=2em\hskip0em\@makefnmark#1}
\makeatother

\begin{document}

\maketitle
\thispagestyle{empty} 

\begin{abstract}
This paper discusses a large-scale and long-term mapping and localization scenario using the maplab open-source framework\footnotemark[1].
We present a brief overview of the specific algorithms in the system that enable building a consistent map from multiple sessions.
We then demonstrate that such a map can be reused even a few months later for efficient 6-DoF localization and also new trajectories can be registered within the existing 3D model.
The datasets presented in this paper are made publicly available.
\end{abstract}

\section{INTRODUCTION AND RELATED WORK}
\vspace{-3pt}

Building, maintaining and localizing against a large-scale map is a key capability for numerous autonomous systems.
Mobile robots depend on a precise 6-DoF pose within the known environment to perform navigation, path planning and interaction.
Robust estimates of the current location and the structure nearby help to act faster and more efficiently. 
A reliable lifelong mapping system paves a way for long-term autonomy of the agents.


However, large-scale maps, collected over long periods of time contain thousands of keyframes and millions of 3D landmarks.
This large number of elements leads to excessive CPU and memory consumption while processing the maps.
Furthermore, it slows down and degrades the performance of localization queries, as the algorithm needs to comb through a large database and often returns outliers.

Accurate vision-based reconstruction at city scale has been demonstrated by the research community, even using images harvested from the web \cite{snavely2006photo}\cite{agarwal2011building}.
Using 3d reconstructions to retrieve a precise 6-DoF pose has been discussed in many works e.g. \emph{Sattler et al.}~\cite{sattler2011fast}.
Such approaches have been adopted by the robotics community and combined with visual-inertial odometry~\cite{mur2015orb} and online map tracking~\cite{lynen2015get}.
Most of the research in visual-inertial estimation focuses on single session mapping at building scale without place recognition capabilities \cite{bloesch2015robust}\cite{leutenegger2015keyframe}.

In \cite{schneider2018maplab}, \emph{Schneider et al.} introduced maplab\footnotemark[1], a research-oriented visual-inertial mapping and localization framework.
The system was evaluated on a large multi-session dataset of Zurich's old town.
In this paper, we would like to extend this evaluation by a long-term scenario.
Overall, the contributions of this paper are as follows:
\begin{itemize}
    \item We survey the specific design of the maplab framework that supports large-scale long-term mapping,
    \item We evaluate the performance of the map building and make the Zurich Old Town dataset publicly available,
    \item Finally, we evaluate the localization of a new recording, collected months later, against the original map.
\end{itemize}

\footnotetext[1]{Maplab and the datasets are available at:\\ 
\url{www.github.com/ethz-asl/maplab}}

\begin{figure}[t]
  \centering
  \includegraphics[width=1.0\columnwidth]{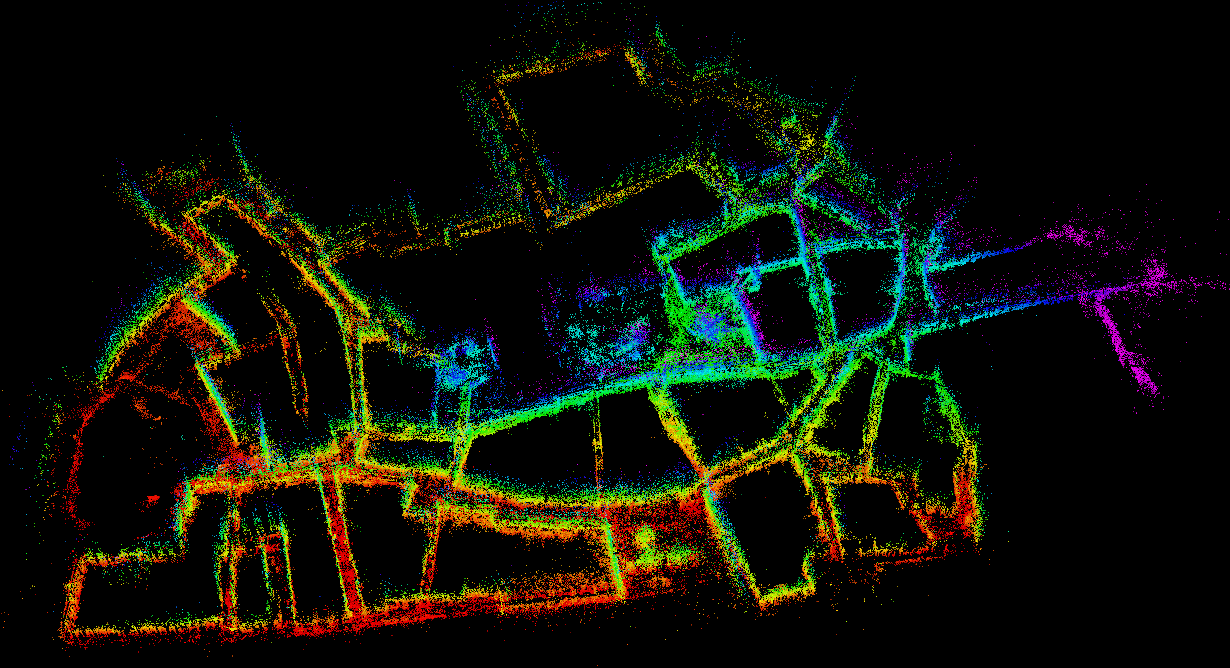}
  \caption{A large-scale reconstruction of the Zurich Old Town dataset.
  maplab outputs consistent reconstructions of large areas out of multiple recordings.
  The processing can be done on a desktop PC and the final result aligns well with the actual city map.
  \vspace{-5mm}}
  \label{fig:teaser}
\end{figure}

\section{METHODOLOGY}
\vspace{-3pt}

This section presents the methodology used by the maplab framework to process and re-use large-scale maps.
In \ref{subsec:build}, we describe the map building stage and introduce the specific adaptations that are targeted at handling multi-session, long-term and large-scale visual-inertial maps.
In \ref{subsec:localize}, we briefly present the descriptor-based localization system used for experiments.
All operations presented below can be executed directly within the maplab console.

\subsection{Building a consistent, large-scale map}
\label{subsec:build}
\vspace{-3pt}

\textbf{Collecting data:} The maps can be recorded using any synchronized visual-inertial sensor, such as a Google Tango Yellowstone tablet or Intel ZR300 sensors.
The data is processed using ROVIOLI in the VIO mode (see \cite{schneider2018maplab}) and loaded as a single map to maplab.

\textbf{Redundant keyframe removal:} The maplab framework offers an option to keyframe the map, i.e. discard frames that contain mostly redundant information with neighbouring frames and are thus not relevant for maintaining a consistent map geometry.
The keyframes are selected based on:
\begin{itemize}
    \item maximum translation and rotation between frames,
    \item maximum number of consecutive removed frames,
    \item and a minimum number of co-observed landmarks.
\end{itemize}

\textbf{Filtering of unreliable landmarks:} The maps are further optimized by detecting possibly unreliable landmarks.
The subsequent operations ignore landmarks with a small number of observers (less than 4), small disparity angle between observations (less than $5\degree$) and a distance from the observer that is excessively large.

\textbf{Efficient rigid alignment of the maps:} The separate trajectories resulting from each recording are then rigidly pre-aligned.
The system uses an efficient loop-detector engine to probe for possible correspondences between each pair of trajectories.
The matches are used to estimate a relative 6-DoF transformation between the trajectories.
The pairwise querying ensures even trajectories with a limited overlap can be correctly anchored in the global frame.

\textbf{Robust posegraph relaxation:} Significant geometric discrepancies between trajectories are rectified using robust posegraph optimization~\cite{sunderhauf2012switchable}.
The corrected geometry improves the convergence of the the subsequent full-batch optimization. 

\textbf{Merging landmarks:} The next step consists of merging the duplicated 3D landmarks.
The loop-detector engine is used to merge corresponding landmarks and  the keyframe-landmark observation backlinks are updated accordingly.

\textbf{Refining the geometry:} After merging the landmarks, the geometry of the map is further refined using a full-batch visual-inertial weighted least-squares optimization.

\textbf{Summarizing the map:} Finally, only the landmarks useful for localization are retained by the map summarization~\cite{dymczyk2015keep}.

\subsection{Localization against the large-scale map}
\label{subsec:localize}
\vspace{-3pt}

Our localization system is based on the principle of the 2D-3D matching~\cite{sattler2011fast}, i.e. identifying the correspondences between the current query frame and the prior 3D model.

\textbf{Descriptor matching:} The matches are established using BRISK descriptors that are projected to an Euclidean space and retrieved using an inverted multi index~\cite{lynen2015get}.
Finally, the matches are filtered based on covisibility information.

\textbf{6-DoF pose estimation:} The matches are then passed to a pose estimation module that consists of a PnP RANSAC.
The consensus method filters outliers and provides an accurate 6-DoF pose.
This pose can be directly used by algorithms such as planning or navigation.

\section{EXPERIMENTAL EVALUATION}
\vspace{-3pt}

\subsection{Building a consistent, large-scale map}
\label{subsec:dbmap}
\vspace{-3pt}

To evaluate the maplab framework, we have recorded more than $16$km of trajectories in 45 sessions covering the Zurich Old Town area.
The recordings were collected over two days.

The maps were then processed on a single PC in less than $10$h.
The full-batch optimization was the most demanding task with the peak memory consumption of 12.9~GB and took $512.3$s per iteration.
The resulting map contains 435k landmarks, 7.3M descriptors, over 21k keyframes and aligns well with the city geometry (see figure \ref{fig:teaser}).

\subsection{Long-term localization}
\vspace{-2pt}

\begin{figure}
  \centering
  \includegraphics[width=1.0\columnwidth]{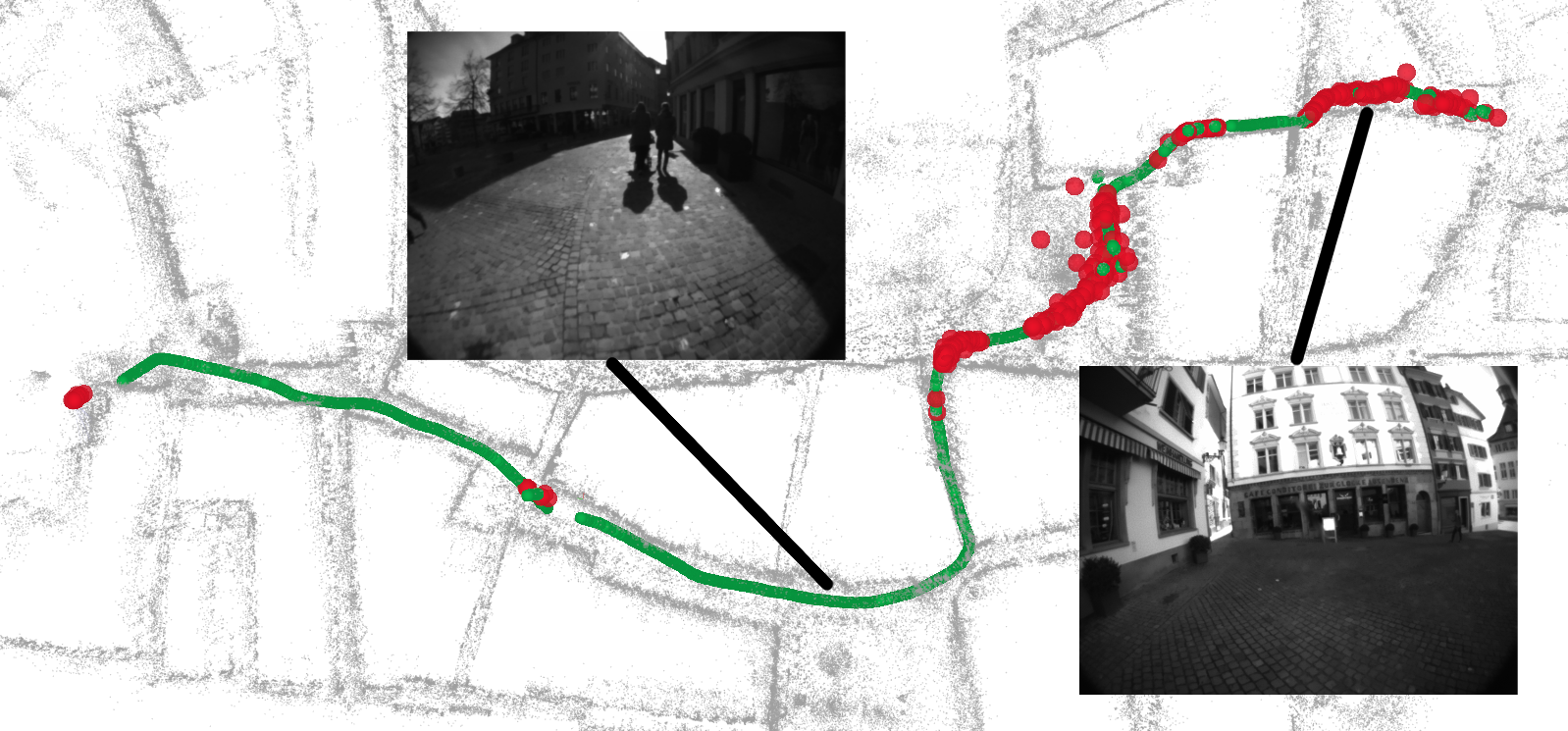}
  \caption{The localization result of a new trajectory recorded 6 months after the collection of the original dataset.
  The green trajectory is an output of ROVIOLI, that fuses visual-inertial data and the 6-DoF localizations.
  The red dots denote raw localization events.
  The localization permits to register the new trajectory and correct for drift.
  A part of the trajectory was not localized which we attribute to poor illumination conditions, i.e. shadows and sun glare, as visible in the left camera image.
  \vspace{-5mm}
  }
  \label{fig:loc}
\end{figure}

To verify the long-term localization performance, about 6 months later we have collected an additional dataset of $521$m length that contains 1796 frames.
Each of the frames was queried against the map (section~\ref{subsec:dbmap}), which required $28$ms on average.
A default loop-closure plugin with default settings was used.
The maplab framework correctly retrieved poses of over 20\% of the frames (362 frames), as presented in figure~\ref{fig:loc}.
No false positive results (10m radius)  were returned.

\section{CONCLUSIONS}
\vspace{-3pt}

This work presents large-scale mapping and localization capabilities of the open-source maplab framework.
We have introduced the specific features that permit to accommodate maps that span over few city blocks.
We have also demonstrated that a long-term localization within a large-scale map is feasible and leads to a correct co-registration.
%

\addtolength{\textheight}{-12cm}   


\vspace{-3pt}
\small
\bibliographystyle{IEEEtran} 
\bibliography{robotvision}
\end{document}